\documentclass[preprint,11pt]{elsarticle}

\usepackage{amssymb}
\usepackage{amsfonts}
\usepackage{nicefrac}

\usepackage{times}
\usepackage{float}
\usepackage{amsmath}
\usepackage{changepage}
\usepackage{graphicx}
\usepackage{caption}
\usepackage{relsize}
\usepackage{times}
\usepackage{subfigure}
\usepackage{natbib}
\usepackage{hyperref}
\usepackage{mathtools}

\usepackage{algorithm}
\usepackage{algorithmic}
\usepackage[section]{placeins}

\DeclarePairedDelimiter\abs{\lvert}{\rvert}%
\makeatletter
\let\oldabs\abs
\def\abs{\@ifstar{\oldabs}{\oldabs*}}
\let\oldnorm\norm
\def\norm{\@ifstar{\oldnorm}{\oldnorm*}}
\makeatother

\usepackage{letltxmacro}
\LetLtxMacro\oldttfamily\ttfamily
\DeclareRobustCommand{\ttfamily}{\oldttfamily\csname ttsize\endcsname}
\newcommand{\setttsize}[1]{\def\ttsize{#1}}%

\usepackage{lineno}

\journal{Automation in Construction}

\makeatletter
\def\ps@pprintTitle{%
 \let\@oddhead\@empty
 \let\@evenhead\@empty
 \def\@oddfoot{}%
 \let\@evenfoot\@oddfoot}
\makeatother
\begin{document}

\setttsize{\small}

\begin{frontmatter}



\title{AI-based Prediction of Independent Construction Safety Outcomes from Universal Attributes \\ {\small\textit{Accepted for publication in Automation in Construction}}}

\author[label1]{Henrietta Baker}
\author[label2]{Matthew R. Hallowell}
\author[label3,label4]{Antoine J.-P. Tixier}
\address[label1]{University of Edinburgh, UK}
\address[label2]{University of Colorado at Boulder, USA}
\address[label3]{\'{E}cole Polytechnique, France}
\address[label4]{corresponding author}

\begin{abstract}
\textbf{Highlights}
\begin{itemize}
    \item This paper significantly improves on and validates ``Application of Machine Learning to Construction Injury Prediction'' (Tixier \textit{et al.} 2016 \cite{tixier2016application}).
    \item NLP is used to extract attributes from a large dataset of injury reports and machine learning models are trained to predict \textit{independent} safety outcomes (human annotations).
    \item Four safety outcomes are predicted: injury severity, injury type, bodypart impacted, and incident type.
    \item Random Forest, XGBoost and linear SVM are compared and combined through model stacking.
    \item Results show that attributes are still highly predictive, even for the \textit{injury severity} outcome. An analysis of per-category attribute importance scores is performed.
\end{itemize}

\noindent\textbf{Abstract}
This paper significantly improves on, and finishes to validate, an approach proposed in previous research in which safety outcomes were predicted from attributes with machine learning.
Like in the original study, we use Natural Language Processing (NLP) to extract fundamental attributes from raw incident reports and machine learning models are trained to predict safety outcomes.
The outcomes predicted here are injury severity, injury type, body part impacted, and incident type.
However, unlike in the original study, safety outcomes were not extracted via NLP but were provided by \textit{independent} human annotations, eliminating any potential source of artificial correlation between predictors and predictands.
Results show that attributes are still highly predictive, confirming the validity of the original approach.
Other improvements brought by the current study include the use of (1) a much larger dataset featuring more than 90,000 reports, (2) two new models, XGBoost and linear SVM (Support Vector Machines), (3) model stacking, (4) a more straightforward experimental setup with more appropriate performance metrics, and (5) an analysis of per-category attribute importance scores. Finally, the \textit{injury severity} outcome is well predicted, which was not the case in the original study. This is a significant advancement.

\end{abstract}


\end{frontmatter}


\section{Introduction}
It is no secret that construction is one of the most dangerous industries. The rate of occupational fatality is around three times the national all-industry average in both the USA and Europe \cite{william2014fatality}.
Predicting injuries has been one of the ultimate goals of construction safety research and practice since the inception of the field. To do so, large quantities of precursors and outcomes must be obtained. Collecting such data is far from trivial: what type of data should be used? How to define precursors? When these questions have been answered, the extraction process is still laborious and time consuming. 

In 2016, \cite{tixier2016application} proposed a novel approach capitalizing on the Natural Language Processing (NLP) system of \cite{tixier2016automated} to automatically extract, at no cost, attributes from raw injury reports.
Attributes are fundamental, context-free descriptors of the work environment that can be observed onsite before an accident occurs.
Additionally, safety outcomes were extracted with the same tool from the free-text narratives.
The authors showed that attributes could predict the outcomes much better than random. However, since both attributes and outcomes were extracted from the same documents, and with the same tool, it could be argued that the algorithms were just capturing some artificial attribute-outcome correlation induced by the NLP method.

In this paper, we address this important limitation, and improve on \cite{tixier2016application} at many other different levels.
More precisely, like in \cite{tixier2016application}, we use the NLP system of \cite{tixier2016automated} to extract fundamental attributes from raw incident reports, and machine learning models to predict safety outcomes. However, the outcomes used in this study were not extracted with the NLP tool. Rather, they were already available in the dataset as \textit{independent} fields that had been filled by safety professionals when each report was entered in the database. In other words, the safety outcomes used in this study are external to the attribute extraction process. This removes any potential source of artificial correlation between predictors and predictands, thereby increasing both the validity and difficulty of the prediction task.

Experimental results with this new configuration show that attributes still predict, with high skill, all independent safety outcomes included in the analysis: \textit{incident type}, \textit{injury type}, \textit{bodypart}, and \textit{injury severity}.
This confirms the validity of \cite{tixier2016application}.
In addition, the fact that the \textit{injury severity} outcome is well predicted too is a significant improvement over \cite{tixier2016application}, who hypothesized it was impossible based on attributes only.
Results also show that the NLP tool of \cite{tixier2016automated} can perform well outside of the domain on which it was originally developed (oil and gas versus industrial, energy, infrastructure, and mining). The main contributions of this study are:

\begin{itemize}
    \item we validate the approach of \cite{tixier2016application} by showing that attributes still have high predictive power when the safety outcomes are external and independent. Also, unlike \cite{tixier2016application}, we predict well even \textit{injury severity}.
    \item We also improve on \cite{tixier2016application} at numerous other levels:
    
    \begin{itemize}
        \item we use a much larger dataset featuring more than 90,000 reports, and experiment with two new state-of-the-art models (XGBoost and linear SVM).
        \item To boost performance, we implement a model stacking strategy.
        \item We adopt a more straightforward experimental setup with more appropriate performance metrics.
        \item Finally, we also explain how per-category attribute importance scores can be obtained and used to understand predictions.
    \end{itemize}

\end{itemize}

To demonstrate the aforelisted contributions, this paper is structured as follows. After giving a brief exploration of the relevant literature (in section \ref{sec:back}) and introducing the attribute framework (section \ref{sec:att_nlp}), we present the dataset (section \ref{sec:data}) and machine learning models used (section \ref{sec:ml}).
Then, we detail our experimental setup (section \ref{sec:exps}) and report and interpret our results (section \ref{sec:res}).
The limitations and implications for further application in construction safety are discussed in section \ref{sec:lims}.
We finally conclude.

\section{Background}\label{sec:back}
\subsection{Machine learning in construction}
Machine learning (ML) is used to describe a wide range of computer programs that implement algorithms and statistical models to carry out tasks \cite{hastie2005elements}.
These programs rely on patterns and inference, rather than explicit instructions, to achieve their aims \cite{bishop2006pattern}.

ML in construction has been developed significantly since 1991 when \cite{moselhi1991neural} first discussed the potential of neural networks in construction engineering and management.
Early examples of ML in construction include applications such as \cite{skibniewski1997constructability} where the AQ15 algorithm was applied to automatically learn the mapping between constructability (poor, good, excellent) and 7 predictors from a collection of 31 training examples; and \cite{soibelman2002data} who applied decision trees and neural networks to a construction management database to identify the causes of delays. 

Many subsequent prediction applications applied support vector machines \linebreak (SVMs), owing to their consistently high accuracy. These applications include \cite{lam2009support}, who accurately forecasted contractor prequalification using input variables such as financial strength and current workload; \cite{cheng2010estimate}, who estimated building cost and loss risk from ten input variables; and \cite{son2011automated}, who detected concrete structural components in color images from actual construction sites.

In the last 5 years, use of ML in construction has become far more widespread and the methods and applications used are far more diverse. In addition to classic prediction tasks, more nuanced applications have emerged. Some interesting examples include construction equipment activity recognition \cite{akhavian2015construction}, and productivity and ergonomic assessment \cite{nath2017construction}.

\subsection{Machine learning for construction safety}
With regard to construction safety, before 1995, research was heavily invested in the analysis of lagging statistics \cite{zhou2015overview}. The aim of such studies, e.g., \cite{hubbard1985major,salminen1995serious}, was to observe trends in accident numbers and postulate correlations with a limited number of circumstantial factors to suggest future safety measures or research avenues. At the same time, statistics concerning safety incidents and their associated cost were used to create financial motivation for safety research, e.g. \cite{koehn1983osha}. Neither of these applications attempted to empirically forecast future trends or safety events, but rather examined the current state and postulated positive actions towards reducing incident rates.\\

Regarding pure prediction of construction safety outcomes from descriptors of the work and the work environment, a recent survey by \cite{hallowell2019methods} recognized two studies \cite{tixier2016application,esmaeili2015attribute}.
Further publications identified in this domain are \cite{kang2019predicting,sarkar2019prediction,sarkar2019sliptripfall}.
However, in all these pieces of work except \cite{tixier2016application}, some of the input variables are outcomes.
Such variables cannot be considered valid predictors as they are not observable before accident occurrence.
E.g., in \cite{esmaeili2015attribute}, \textit{structure collapse} and \textit{falling from roof}, two outcomes, are used as attributes.
The attributes of \cite{sarkar2019prediction,sarkar2019sliptripfall} also include two outcomes: incident type and injury type.
Finally, \cite{kang2019predicting} rely on accident type and injury type too, but also on body part injured and accident location. All of these variables, again, are outcomes, not predictors.

Another relevant study is \cite{poh2018safety}. In this work, the authors rely on 13 project management and safety-related leading indicators from monthly inspection data (before incident occurrence) to make severity forecasts.
While this approach is valid and interesting, it cannot be directly compared to the present study, as using leading indicators and fundamental attributes (as defined in section \ref{sec:att_nlp}) are two different approaches.

Additionally, there are large differences in terms of prediction complexity and methodology.
For example, in \cite{tixier2016application} and in the present study, multiple injury characteristics are predicted using ensemble machine learning methods, including 7 (resp., 4) types of injury, 5 (6) body parts and 6 (2) severity levels. In the present study, we also predict 6 incident types and experiment with model stacking.
Meanwhile, \cite{esmaeili2015attribute} predict only a binary severity outcome (fatal vs. non-fatal) with a simple logistic regression model and \cite{poh2018safety} predict only three categories (no accident, minor accident or major accident) using five standalone machine learning algorithms.


\section{Attribute framework and NLP method}\label{sec:att_nlp}

The attribute-based framework developed by \cite{desvignes2014requisite,villanova2014attribute} allows any construction situation to be described by a finite number of attributes that are observable \textit{before} accident occurrence. 
Attributes span construction means and methods, and environmental conditions. They are universal, basic descriptors of the construction work.

To illustrate, in the following excerpt of an injury report: ``\textit{worker tripped on a cable when carrying a 2x4 to his truck}'', 4 fundamental attributes can be identified: (1) cable, (2) object on the floor, (3) lumber, and (4) light vehicle. It is interesting to note that a simple keyword search would not have identified these attributes while the rule-based NLP model has correctly identified: (2) object on floor attribute by implication from the tripping action; (3) lumber from its slang reference (``2x4''); and included application of expert judgement to realize that ``truck'' refers to a pick-up truck, (4) light vehicle, rather than a heavy lorry. 

The creation of the framework by \cite{desvignes2014requisite,villanova2014attribute} involved manual inspections of 1000+ accident reports by a team of 8 coders over the course of several months with weekly peer reviews, calibration meetings, and voting.
Once the final list of 80 attributes was obtained (see Table \ref{table:atts}), \cite{tixier2016automated} developed a NLP method to automatically extract all of these attributes (and several outcomes) from future reports based on hand-written logical rules and dictionaries of general and specific keywords.
The accuracy of the computer program was verified against external coding by numerous manual reviewers and after several weeks of iterations and fine-tuning, a 95\% agreement threshold was achieved. The care taken in this previous work ensured strong \textit{feature engineering}, which is of paramount importance to ML success.

Note that by allowing attributes and outcomes to be extracted from unlimited amounts of narratives, the NLP program enabled not only the training of predictive models but also large-scale data mining, and risk analysis and simulation \cite{tixier2017construction}.\\

\begin{table}[h]
\small
\centering
\scalebox{0.8}{
\begin{tabular}{llll}
  \hline
adverse low temps & grinding & machinery & spark\\ 
bolt & object at height & manlift & splinter sliver\\ 
cable & guardrail handrail & mud & steel steel sections\\ 
cable tray & hammer & nail & poor visibility \\ 
chipping & hand size pieces & improper PPE & spool\\ 
cleaning & hazardous substance & grout & stripping\\ 
concrete & heat source & object on the floor & stud\\ 
concrete liquid & heavy material/tool & piping & tank\\ 
conduit & heavy vehicle & pontoon & uneven surface\\ 
confined work space & hose & poor housekeeping & insect\\ 
congested work space & imp. body position & stairs & wind\\ 
crane & imp. procedure inattention & powered tool & wire\\ 
door & imp. security of materials & repetitive motion & valve\\ 
drill & imp. security of tools & rebar & welding\\
dunnage & unpowered tool & scaffold & unpowered transporter\\ 
electricity & job trailer & screw & unstable support surface\\ 
exiting & ladder & sharp edge & working below elev wksp mat\\ 
fatigued dizzy & lifting pulling manipulating & slag & wrench\\ 
forklift & light vehicle & slippery surface & working overhead\\ 
formwork & lumber & small particle & working at height \\
\hline
\end{tabular}

}
\captionsetup{size=footnotesize,justification=centering}
\caption{Final list of 80 validated attributes from \cite{desvignes2014requisite,villanova2014attribute} used in this study.}
\label{table:atts}
\end{table}

\noindent \textbf{Attribute extraction + machine learning vs. deep learning for NLP}. Deep learning models are very powerful, as they learn a direct mapping between raw text and outcomes \cite{tixier2018notes}.
They were recently applied to construction injury reports by \cite{baker2019automatically}, with the purpose of automatically learning injury precursors as a side effect of classifying the reports into safety outcome classes.

While the approach of \cite{baker2019automatically} is innovative and offers many advantages, a current limitation is that DL models sometimes use parts of the narratives explicitly discussing outcomes rather than the incident circumstances and environmental conditions.
That is, models have a tendency to ``cheat'' and use the solutions, which reduces their ability to extract meaningful precursors. \cite{baker2019automatically} proposed several solutions to address this problem in future work.

On the other hand, machine learning involves a two-step process: first, a set of features has to be defined (\textit{feature engineering}) and extracted from the input documents (\textit{feature extraction/recognition}), second, a supervised model is trained to predict the outcomes from the features. In this study, the features were the set of 80 fundamental attributes shown in Table \ref{table:atts}.

The two-step ML approach offers some advantages over deep learning: (1) \textit{interpretability}, as attributes were created by human experts and make physical sense, and (2) \textit{validity}, as attributes are observable in the environment \textit{before} incident occurrence. However, some information is lost compared to using the raw text as input, as (1) attributes cannot represent everything about the work environment, and (2) some errors are made during the extraction process (false alarms and misses).

\section{Dataset}\label{sec:data}

\subsection{Text and outcomes}
We had access to a large dataset of more than 90,000 incident reports provided by a global industrial partner in the oil and gas sector. The reports corresponded to work done on five continents from the early 2000s to 2018.

The following outcomes were available for some of the injury reports: 
\begin{enumerate}
    \item  \texttt{severity}: severity of the injury. Severity of the injury was measured by the level of immediate medical care required.
    \item \texttt{injury\_type}: nature of the injury. The four types of injury included in this analysis were \texttt{contusion}, \texttt{cut/puncture}, \texttt{FOB} (foreign object in body) and \texttt{pain}.
    \item \texttt{bodypart}: body part(s) impacted during the incident.
    \item \texttt{incident\_type}: nature of the incident. This category describes the immediate cause of the incident. For example, if the accident occurred due to faulty or misuse of equipment or tools, the class \texttt{eq./tools} would be selected. Likewise, \texttt{rules} refers to rule violation, \texttt{access} to issues with access, \texttt{PPE} to insufficient PPE, and \texttt{dropped} to an injury resulting from a dropped item.
    \end{enumerate} 
Each outcome was associated with a certain number of unique values, that we will denote in the remainder of this articles interchangeably as categories, classes, or levels. Category counts for each outcome are shown in Table \ref{table:out_counts}.

\begin{table}[h]
\centering
\scalebox{0.8}{

\begin{tabular}[t]{lr}
\multicolumn{2}{c}{incident type} \\
\hline
eq./tools    & 29033 \\
slips/trips/falls & 20474 \\
rules         & 14296 \\
access        & 10368 \\
PPE           &  8998 \\
dropped           &  8469 \\
\end{tabular}

\begin{tabular}[t]{lr}
\multicolumn{2}{c}{injury type} \\ 
\hline
contusion     &   4048 \\
cut/puncture &   3737 \\
FOB         &   2073 \\
pain          &   1587 \\
\end{tabular}

\begin{tabular}[t]{lr}
\multicolumn{2}{c}{bodypart} \\
\hline
finger &   4421 \\
hand &   3161 \\
eye    &   2392 \\
lower extr.  &   2321 \\
head     &   2184 \\
upper extr.  &   1757 \\
\end{tabular}

\begin{tabular}[t]{lr}
\multicolumn{2}{c}{severity} \\
\hline
1st aid &  16510 \\
med./restr.  &   2875 \\
\end{tabular}
}
\captionsetup{size=footnotesize}
\caption{Category counts for each outcome. \label{table:out_counts}}
\end{table}

\subsection{Attribute extraction with Natural Language Processing}
We extracted attributes from each report used in this study with the NLP method of \cite{tixier2016automated}, previously described. Note that this method was developed and fine-tuned on industrial, energy, infrastructure, and mining reports. However here, we apply it to reports belonging to the oil and gas sector.

\subsection{Train/test splits and class imbalance} \label{subsec:traintest}
We created separate train/test splits for each outcome: 90\% of the reports were used for training, and 10\% were left-out for testing. For parameter optimization purposes, a validation set was also created for each outcome by randomly setting aside 10\% of the training set.

Categories were not mutually exclusive for \texttt{injury\_type} and \texttt{bodypart}. For these outcomes, 301 and 519 cases (respectively) were associated with more than one level. In order to make use of as much data as possible, we considered these cases to belong to \textit{all} the categories they were associated with, rather than arbitrarily selecting a single one. For instance, a case associated with `head' and `upper extremities' was used as training data for both `head' and `upper extremities'. However, to make sure that no report from the training sets leaked to the test sets, such extension was performed only after the initial train/test splits had been made. 

To address the problem of class imbalance, weights inversely proportional to the frequency of each category in the training set were computed. These weights were used to make sure that the ML models did not neglect the minority categories. An example of the statistics about the final train/test splits is shown in Table \ref{table:tt_splits_1}.

\begin{table}[h]
\centering
\setlength{\tabcolsep}{2pt}
\scalebox{0.7}{
\begin{tabular}[]{rcccccc|cccccc}
 & \multicolumn{6}{c|}{train (82,474)} 
& \multicolumn{6}{c}{test (9,164)} \\
\hline
 & access & eq./tools & slips/trips/falls & dropped & PPE & rules & access & eq./tools & slips/trips/falls & dropped & PPE & rules \\ 
 counts & 9,300 & 26,167 & 18,432 & 7,619 & 8,078 & 12,878 & 2,866 & 850 & 1,068 & 2,042 & 920 & 1,418 \\ 
 weights & 2.8 & 1.0 & 1.4 & 3.4 & 3.2 & 2.0  &  - & -  & -  & -  & -  & -  
\end{tabular}
}
\captionsetup{size=footnotesize}
\caption[Train/test splits statistics for \texttt{incident\_type}]{Train/test splits statistics for \texttt{incident\_type}.\label{table:tt_splits_1}}
\end{table}

\section{Machine learning models}\label{sec:ml}

We experimented with three state-of-the-art machine learning models: Random Forest \cite{breiman2001random}, extreme gradient boosting (XGBoost) \cite{chen2016xgboost}, and linear Support Vector Machine (SVM) \cite{boser1992training}. All are widely used in practice and are among the most successful machine learning algorithms to date.

More precisely, we capitalized on the Python's \texttt{scikit-learn} implementations of Random Forest\footnote{\href{https://scikit-learn.org/stable/modules/generated/sklearn.ensemble.RandomForestClassifier.html}{https://scikit-learn.org/stable/modules/generated/sklearn.ensemble.RandomForestClassifier.html}} and linear SVM\footnote{\href{https://scikit-learn.org/stable/modules/generated/sklearn.svm.LinearSVC.html}{https://scikit-learn.org/stable/modules/generated/sklearn.svm.LinearSVC.html}}, while, for XGBoost, we used the original library\footnote{\href{https://xgboost.readthedocs.io/en/latest/python/python_api.html\#module-xgboost.sklearn}{https://xgboost.readthedocs.io/en/latest/python/python\_api.html\#module-xgboost.sklearn}} (also in Python).

In what follows, we provide details about each algorithm. While they differ in the way they combine trees, RF and XGB are both ensemble learning techniques and are both made of large numbers of decision trees grown with the Classification And Regression Trees (CART) algorithm \cite{breiman1984classification}. Therefore, we first introduce CART before presenting Random Forest and XGBoost. Then, we present the linear SVM algorithm. Also, note that we present each model within the context of \textit{classification}, which is the task of interest in this study.

\subsection{CART}
\begin{figure}[h]
\centering
\includegraphics[width=0.95\textwidth]{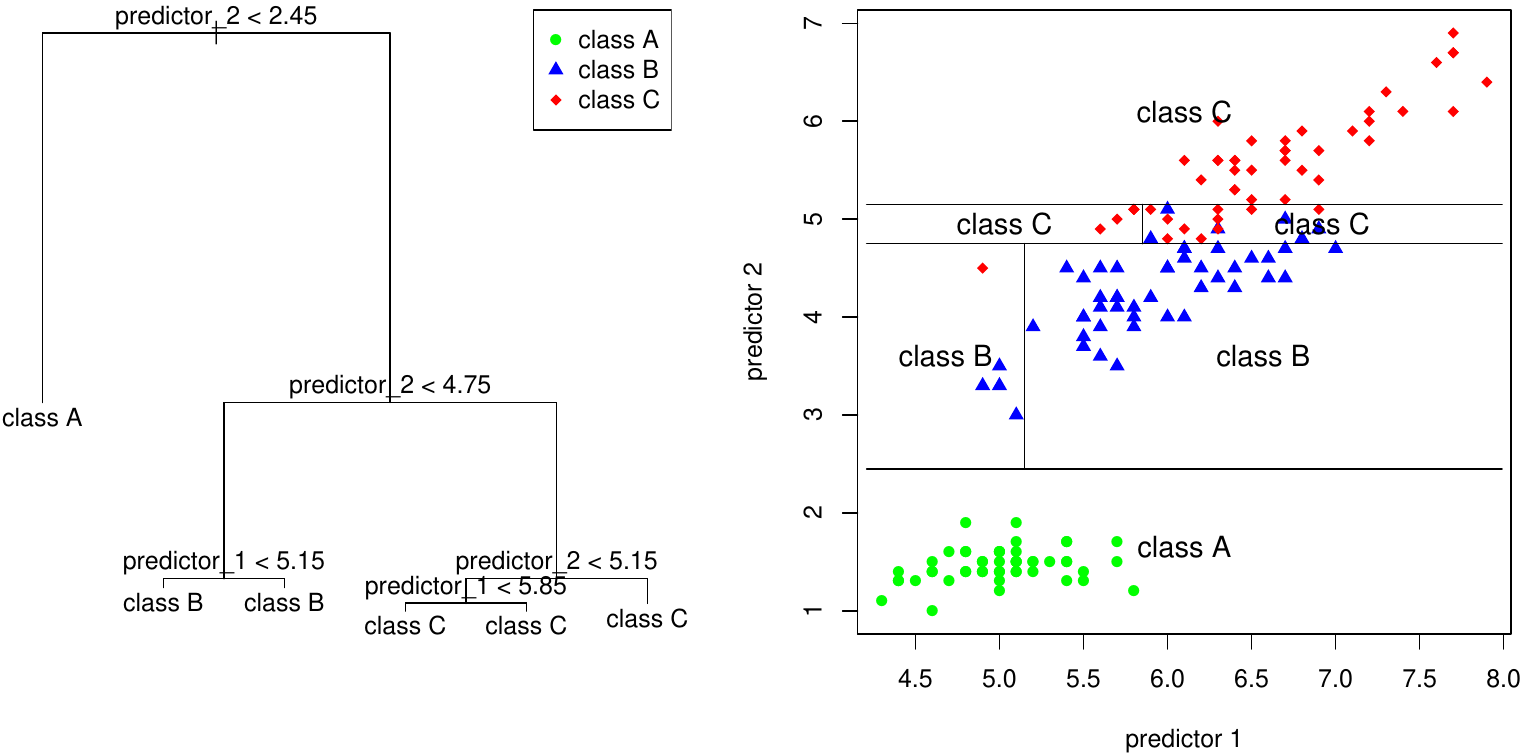}
\captionsetup{size=footnotesize}
\caption{Decision tree toy example. For visualization purposes, the two predictors are continuous. \label{fig:cart}}
\end{figure}

As shown in Algorithm \ref{alg:cart} and Fig. \ref{fig:cart}, CART is a greedy algorithm whose goal is to learn a set of binary recursive rules that partition observations living in a multidimensional feature space. In our case, the dimensions of the space are the 80 fundamental construction attributes.

\begin{algorithm}[h]
\caption{CART\label{alg:cart}}
\begin{algorithmic}[1]
\REQUIRE dataset of observations and features, target variable
\ENSURE set of partitioning rules in the form of recursive binary splits
\STATE start with a root region that includes all observations
\STATE for each predictor, compute a `goodness of split' value based on a criterion
\STATE pick the best predictor (if multiple predictors are best, select the first one)
\STATE split the region in two based on the values taken by the observations on the predictor selected
\STATE repeat steps 2-4 for each new region until the stopping criterion is reached
\end{algorithmic}
\end{algorithm}

More precisely, the final partitioning should be such that most of the observations in the same subspace (or leaf) belong to the same category.
The most frequent category of the observations in a given leaf is then used as the prediction for these observations, as shown in Fig. \ref{fig:cart}. As opposed to \textit{global} models such as linear regression, where the same equation holds over the entire space, decision trees are \textit{local} models.

\noindent For classification, the criterion used at step 2 in Algorithm \ref{alg:cart} was the decrease in Gini diversity index, and the predictor that was selected at step 3 was the one that was maximizing the decrease in Gini index. 
The Gini index measures the heterogeneity, or impurity, of a partition. For instance, it considers a partition 100\% pure if it contains only observations belonging to the same class. The stopping criterion at step 5 was when a certain depth had been reached.

\noindent CART decision trees are able to capture complex nonlinear high-order interactions among predictors, scale well with the number of predictors and observations, and are relatively robust to outliers and irrelevant predictors. However, they often need to be grown very large to accurately represent the training data. This has two negative side effects: (1) poor generalization to unseen observations (overfitting) and (2) high variance, as the lower parts of the trees are very sensitive to changes in the training data. From the perspective of the bias-variance framework, where $error = bias + variance$, deep decision trees are \textit{\textbf{low bias-high variance}} models.

\subsection{Random Forest}\label{subsub:rf}
\noindent\textbf{Bagging}. The \textbf{B}ootstrap \textbf{agg}regat\textbf{ing} (bagging) method \cite{breiman1996bagging} was introduced as a way to take advantage of the low bias of deep decision trees while reducing their high variance. The bagging procedure consists in training many deep trees in parallel on bootstrap samples of the data. A bootstrap sample is obtained by randomly selecting observations with replacement from the original training set until a dataset of the same size is obtained. 
Approximately one third of the observations are not expected to be present in each bootstrap sample, as the probability of not selecting a given observation with replacement from a sample of size $n$ is $\big(1 -\nicefrac{1}{n}\big)^n$, which tends to $\mathrm{exp}(-1) \approx \nicefrac{1}{3}$ when $n$ tends to infinity. These observations compose what is called the `out-of-bag' (OOB) sample \cite{breiman1996out}. Since the bootstrap sample is of same size as the original dataset, it follows that for a large number of observations, each bootstrap sample is expected to contain about two thirds of unique examples, the rest being duplicates. This causes each tree in the ensemble to become an expert on some specific domains of the training set. Bagging thus creates an \textbf{\textit{ensemble of local experts}}.

\begin{figure}[h]
\centering
\includegraphics[width=0.95\textwidth]{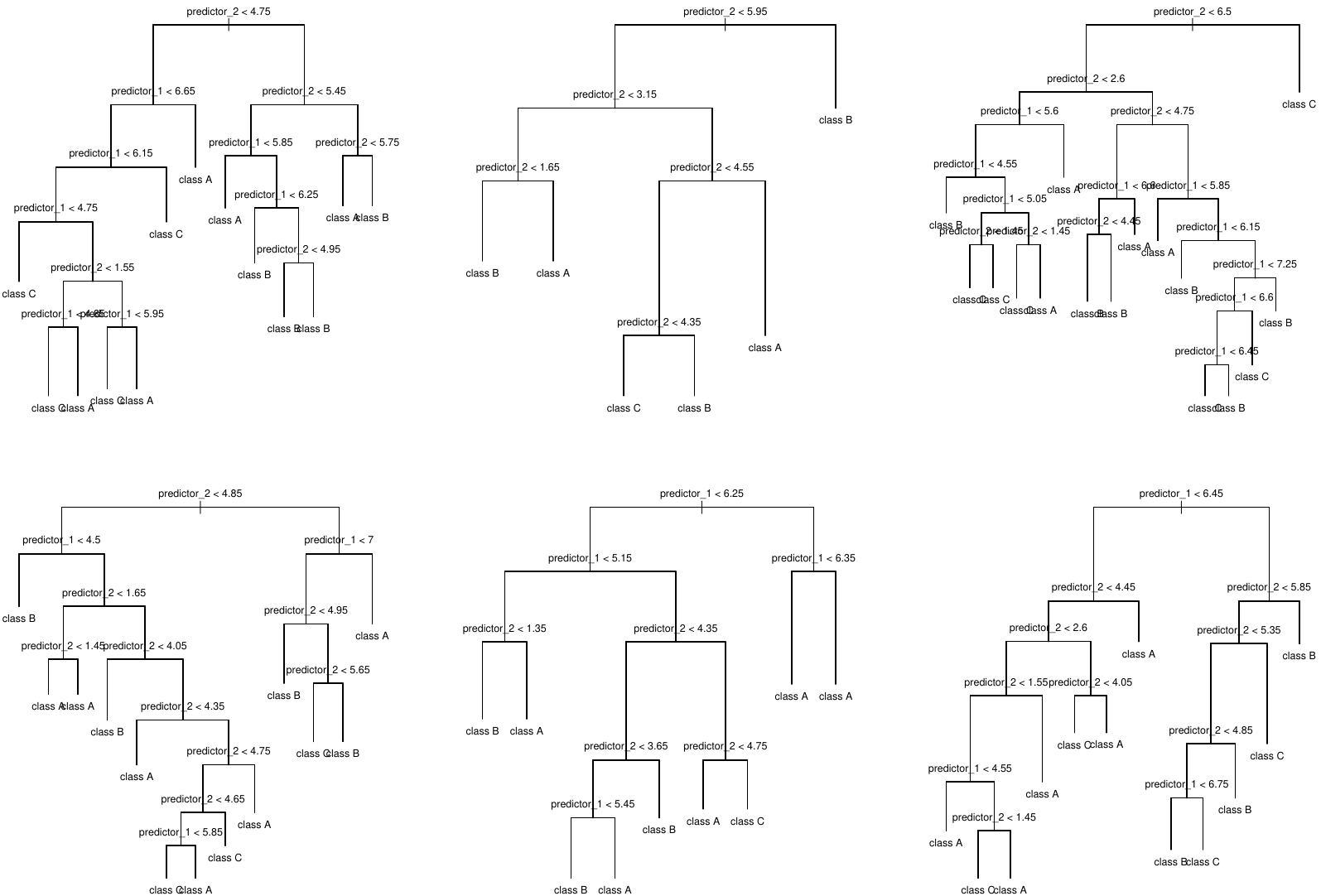}
\captionsetup{size=footnotesize}
\caption{Example of a bagged ensemble of decision trees, using the same data as in Fig. \ref{fig:cart}. Each tree in the ensemble is grown on a bootstrap sample of the original data. The large differences in the tree structures highlight well the high-variance nature of decision trees. \label{fig:bagging}}
\end{figure}

\noindent Thus, at prediction time, there will be a significant amount of beneficial disagreement among trees (see Fig. \ref{fig:bagging}). By aggregating the predictions of all trees in the ensemble via majority voting, one obtains a model with significantly less variance than a single tree. Such a model generalizes much better, while still having almost the same low bias. This approach is known as \textbf{\textit{perturb and combine}}.

Despite being a significant improvement over CART, bagged ensembles are less interpretable. Also, by definition of CART, only those variables yielding the greatest decrease in node impurity are selected at each split. Consequently, all the trees in the bagged ensemble have quite similar upper structures, and tend to generate correlated forecasts, which reduces the disagreement among trees and prevents the maximal reduction in variance from being achieved. \\

\noindent\textbf{Random Forest}. Random Forest \cite{breiman2001random} was designed to address the correlation problem previously described. In Random Forest, trees are still grown on bootstrap samples of the training set, like in bagging. However, a simple modification of the CART algorithm is used: instead of trying \textit{all} predictors at each split, only a \textit{random subset} of them is considered. In practice, this extra randomization gives all predictors a chance to play a role in determining the upper structure of trees, which introduces a lot of beneficial variety in the ensemble and results in greater variance reduction, smaller error rates, and more accurate predictions than with bagging.

\subsection{XGBoost}
Like Random Forest, the boosting algorithm \cite{freund1997decision} is an ensemble approach that combines many base models and let them vote to generate forecasts. However, this apparent similarity is misleading, since Random Forest and boosting tackle the task of error reduction in opposite ways. Indeed, while Random Forest seeks to reduce error by \textit{decreasing the variance} of complex low bias-high variance base models (deep decision trees) built in \textit{parallel}, boosting achieves the same goal by adding weak high bias-low variance base models (shallow decision trees) in \textit{sequence}, repeatedly \textit{reducing the bias} of the entire sequence.

Each shallow tree is a weak learner, i.e., only slightly better than random guessing, but by adding them such that each successive tree is trained to predict the observations that were missed by the preceding one, boosting creates a strong learner. More precisely, at each step, each new tree is fed the pseudo-residuals given by the \textit{gradient} of some differentiable loss function with respect to the predictions of the current model, that is, the entire sequence of trees thus far. This approach is known as \textbf{gradient boosting}.

Moreover, it was empirically shown that training each tree on a \textit{random} subsample of the training set (instead of the full training set) was very beneficial. This method is named \textbf{stochastic gradient boosting}, to emphasize the instillation of randomness into the procedure.

Finally, \textbf{extreme gradient boosting} (XGBoost) \cite{chen2016xgboost} adds a regularization term to the loss function of SGB in order to penalize the complexity of the model, and implements a number of optimization tricks to speed-up training.

\subsection{Linear SVM}\label{subsub:svm}
\begin{figure}[h]
\centering
\includegraphics[width=0.45\textwidth]{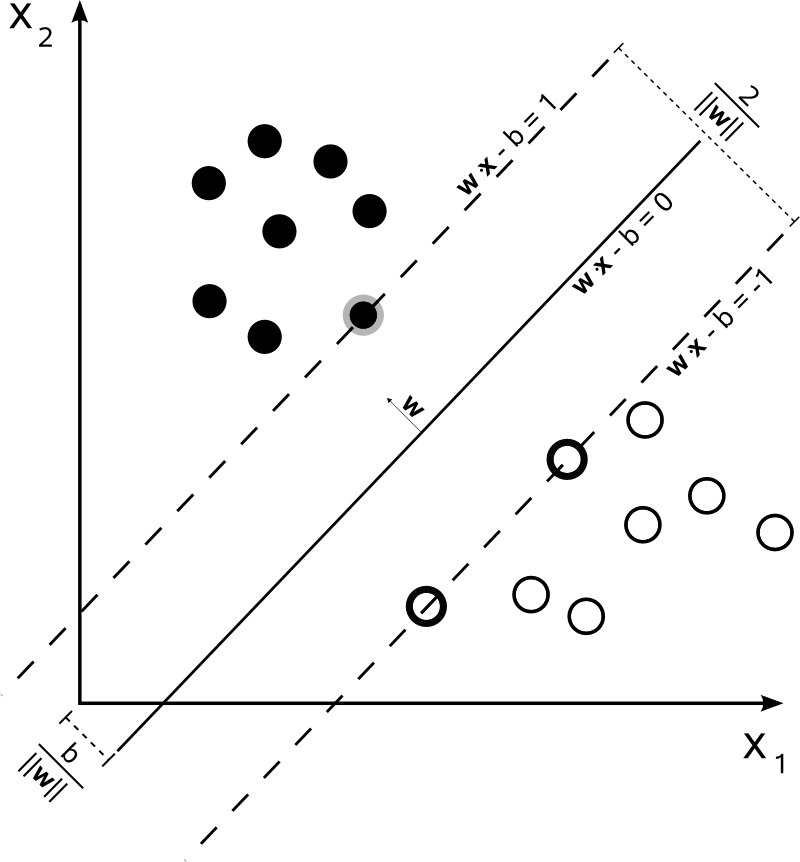}
\captionsetup{size=footnotesize}
\caption{Linear SVM decision boundary in the two-dimensional case (two attributes). \label{fig:svm}}
\end{figure}

The linear Support Vector Machine \cite{boser1992training} is a geometrical method that seeks to classify points into two different categories by finding the best separating hyperplane. As illustrated by Fig. \ref{fig:svm} in the two-dimensional case (two attributes), the best separating hyperplane is the line that separates the two groups of points with the greatest possible margin on each side. Training the SVM, that is, finding the best hyperplane, comes down to optimizing the $\vec{w}$ and b parameters (see Fig. \ref{fig:svm}). In test mode, a new observation is classified based on the side of the hyperplane on which it falls, which corresponds to the sign of the dot product of the observation with the vector orthogonal to the hyperplane ($\vec{w}$).

\noindent Because in practice, points may not all be  separable (e.g., due to outliers), when searching for the best separating hyperplane, the SVM is allowed to misclassify certain points. The tolerance level is controlled by a parameter traditionally referred to as C in the literature. The smaller C, the more tolerant the model is towards misclassification.

C plays a crucial \textit{regularization} role, i.e., it has a strong impact on the generalization ability of the SVM. Indeed, for large values of C (low misclassification tolerance), a smaller-margin hyperplane will be favored over a larger-margin hyperplane if the former classifies more points correctly, at the risk of overfitting the training data. On the other hand, small values of C will favor larger-margin separating hyperplanes, even if they misclassify more points. Such solutions tend to generalize better.

When the target variable features more than two categories, a \textit{one-versus-rest} approach is used to redefine the problem as a set of binary classification tasks. More precisely, as many SVMs as there are categories are trained, and the goal of each SVM is to predict whether an observation belongs to its associated category or not.

\noindent \textbf{Time complexity}. 
Finding the support vectors scales quadratically with the number of training examples $n$. More precisely, it has $\mathcal{O}(n^2)$ complexity when the $C$ parameter is small and $\mathcal{O}(n^3)$ when it gets large \cite{bottou2007support}. In practice, time complexity is the main limitation of SVMs compared to RF or XGBoost. On some large datasets, using a SVM can just be intractable.

\subsection{Attribute importance measures}\label{subsub:attimp}

\noindent \textbf{Random Forest}. The out-of-bag (OOB) observations (see subsection \ref{subsub:rf}) can be used to compute a relative importance score for each predictor. For a given predictor and tree, the procedure consists in randomly permuting the values of the predictor in the set of observations that have not taken part in the training of the tree (i.e., the OOB sample), and comparing the prediction error of the tree on the permuted OOB sample with the prediction error of the tree on the untouched OOB sample. This process is repeated for all the trees in the forest, and the predictor is given an importance score proportional to the overall increase in error that its permutation induced. The most important variables are the ones leading to the greatest loss in predictive accuracy when noised-up \cite{breiman2001random}.\\

\noindent \textbf{XGBoost}. Like Random Forest, the boosting algorithm allows the calculation of importance scores for the predictor variables. In our classification case, the procedure is as follows: for a given tree in the sequence, and for a given non-terminal node of this tree, the reduction in node purity weighted by the number of observations in the node is attributed to the predictor the split was made on. This process is repeated for every non-terminal node of the tree, and the variable importance scores are averaged over all trees.\\

\noindent \textbf{Linear SVM}. Recall from subsection \ref{subsub:svm} that the $\vec{w}$ vector, orthogonal to the best-separating hyperplane, is used to determine whether a given observation belongs to the class of interest or to any of the other classes (one-vs-rest approach). Since by definition, observations can only have non-negative coordinates in our attribute space (0 or 1), a given observation belongs to the class of interest if its dot product with $\vec{w}$ is positive.

Furthermore, the $\vec{w}$ vector contains the contribution of each attribute in making the classification decision. More precisely, the magnitudes of the coordinates of $\vec{w}$ indicate the strength of the contributions, while their signs indicate if the attributes attract or reject observations to/from the class of interest. Attributes for which the coefficients of $\vec{w}$ are large and positive (resp. negative) are strongly indicative of belonging (resp. not belonging) to the category of interest.

\subsection{Model stacking}
To see whether performance could be further improved, we experimented with \textit{stacking}, a popular technique among machine learning practitioners. With stacking, the goal is to automatically learn how to combine the individual strengths of each model with a meta-model, in order to obtain a better classifier. We used simple logistic regression models\footnote{\href{https://scikit-learn.org/stable/modules/generated/sklearn.linear_model.LogisticRegression.html}{https://scikit-learn.org/stable/modules/generated/sklearn.linear\_model.LogisticRegression.html}} as our meta-models, with $C=0.2$ for all outcomes (no tuning was performed). More precisely, a given logistic regression model was trained to predict the levels of a given outcome based on the sum of the probabilistic forecasts of XGB, RF, and SVM.\\

\noindent \textbf{SVM issue}.
By design, the \texttt{linearSVC} implementation of the linear SVM model only returns discrete predictions, that is, a single label corresponding to the most likely class, rather than a probability distribution over \textit{all} classes. Therefore, we converted the discrete predictions into one-hot vectors in order to allow summation with the probabilistic forecasts of XGB and RF. But this approach did not give good results, as adding the SVM to any stack (RF, XGB, RF+XGB) did not outperform any single model, for all outcomes.

Since there is no reliable way to make \texttt{linearSVC} output class probabilities instead of a discrete prediction, we re-trained the best SVM models using the more general \texttt{SVC} implementation\footnote{ \href{https://scikit-learn.org/stable/modules/generated/sklearn.svm.SVC.html}{https://scikit-learn.org/stable/modules/generated/sklearn.svm.SVC.html}}, specifying a linear Kernel. However, while \texttt{linearSVC} is very fast as it is based on the \texttt{LIBLINEAR} library \cite{fan2008liblinear}, \texttt{SVC} is based on the \texttt{LIBSVM} library \cite{chang2011libsvm} and as such does not scale to datasets containing more than 10,000 observations in practice\footnote{\href{https://stackoverflow.com/questions/11508788/whats-the-difference-between-libsvm-and-liblinear}{https://stackoverflow.com/questions/11508788/whats-the-difference-between-libsvm-and-liblinear}}. This made it impossible to include SVM in our stacking experiments for the \texttt{incident\_type} outcome, for which there are more than 80,000 training examples.

Finally, even with probabilistic forecasts, adding the SVM to any stack was decreasing performance, for all outcomes. One possible explanation is that unlike RF or XGBoost, SVMs do not natively return probability estimates. In \texttt{SVC}, Platt's scaling is employed (see section 7.1 of \cite{wu2004probability}). The probabilities obtained with such post-processing methods may not be calibrated in a way that make them easily combinable with other the probabilistic forecasts of other classifiers.

Therefore, we only included XGB and RF in our stacks, for all outcomes.

\section{Experimental setup}\label{sec:exps}

\subsection{Hyperparameter optimization}
We tuned hyperparameters by performing grid searches on the validation sets. The best parameter values for each model and each outcome are shown in Tables \ref{table:best_params_xgb} to \ref{table:best_params_svm}.
More precisely, for Random Forest\footnote{\href{https://scikit-learn.org/stable/modules/generated/sklearn.ensemble.RandomForestClassifier.html}{https://scikit-learn.org/stable/modules/generated/sklearn.ensemble.RandomForestClassifier.html}}, we searched the number of trees (\texttt{ntree} parameter, from 100 to 1200 with steps of 100), the number of variables to try when making each split (\texttt{mtry}, from 5 to 45 with steps of 5), and the leaf size (\texttt{nodesize}, 1, 2, 5, and 10).
For XGBoost\footnote{ \href{https://xgboost.readthedocs.io/en/latest/parameter.html}{https://xgboost.readthedocs.io/en/latest/parameter.html}}, we searched the maximum depth of a tree in the sequence (\texttt{max\_depth}, from 3 to 6 with steps of 1), the learning rate (\texttt{learning\_rate}, 0.01, 0.05, and 0.1), the minimum leaf size (\texttt{min\_child\_weight}, 1,3, and 5), the percentage of training instances to be used in building each tree (\texttt{subsample}, 0.3, 0.5, 0.7, and 1) , and the percentage of predictors to be considered in making each split of a given tree (\texttt{colsample\_bylevel}, 0.3, 0.5, 0.7, and 1). The number of trees in the sequence (\texttt{ntrees}) was automatically selected with an early stopping strategy, where the validation loss had to decrease at least once every 200 iterations to continue training. The loss was the multinomial or binary log loss, depending on the dataset. Note that the maximum allowed number of trees in any case was 2000.
Finally, for the SVM model, we optimized the C parameter (\texttt{C}, $10^x$ with $x$ taking 800 evenly spaced values in $[-7,7]$).

\begin{table}[H]
\centering
\scalebox{0.6}{
\begin{tabular}{rrrrrrr}

 & {\small\texttt{max\_depth}} & {\small\texttt{learning\_rate}} & {\small\texttt{min\_child\_weight}} & {\small\texttt{subsample}} & {\small\texttt{colsample\_bylevel}} & {\small\texttt{ntrees}} \\ 
  \hline
injury severity & 5 & 0.10 & 5 & 0.7 & 1.0 & 326 \\ 
  body part & 6 & 0.10 & 1 & 0.5 & 0.5 & 105 \\ 
  injury type & 3 & 0.10 & 5 & 1.0 & 0.3 & 662 \\ 
  case categorisation & 3 & 0.05 & 1 & 1.0 & 1.0 & 1845 \\ 

\end{tabular}
}
\captionsetup{size=footnotesize}
\caption[Best hyperparameter values for XGBoost]{Best hyperparameter values for XGBoost. \label{table:best_params_xgb}}
\end{table}

\vspace{-0.6cm}

\begin{table}[H]
\centering
\scalebox{0.7}{
\begin{tabular}{rrrr}

 & {\small\texttt{ntree}} & {\small\texttt{mtry}} & {\small\texttt{nodesize}} \\ 
  \hline
injury severity & 700 &  30 &   2 \\ 
  body part & 800 &  20 &   5 \\ 
  injury type & 600 &  15 &   5 \\ 
  case categorisation & 600 &  10 &   2 \\ 

\end{tabular}
}
\captionsetup{size=footnotesize}
\caption[Best hyperparameter values for Random Forest]{Best hyperparameter values for Random Forest. \label{table:best_params_rf}}
\end{table}

\vspace{-0.6cm}

\begin{table}[H]
\centering
\scalebox{0.7}{
\begin{tabular}{rr}

 & {\small\texttt{x}} \\ 
  \hline
injury severity & -1.638 \\ 
  body part & -0.920 \\ 
  injury type & -1.778 \\ 
  case categorisation & -1.253 \\ 

\end{tabular}
}
\captionsetup{size=footnotesize}
\caption[Best hyperparameter values for SVM]{Best hyperparameter values for linear SVM, where $C=10^{x}$ \label{table:best_params_svm}}
\end{table}

The final models, corresponding to the best parameter combinations, were then trained on the union of the training and validation sets and tested on the test set. To alleviate the negative effects of class imbalance, weights inversely proportional to the size of each category were used (category weights for RF and SVM, sample weights for XGB).

\subsection{Configuration}
For the XGBoost grid searches, we used the GPU-accelerated implementation of the fast histogram algorithm (\texttt{gpu\_hist}) as the tree method\footnote{\href{https://xgboost.readthedocs.io/en/latest/gpu/}{https://xgboost.readthedocs.io/en/latest/gpu/}}, which led to a significant 8X speedup. However, we made sure to use the exact algorithm (\texttt{exact}) for training and testing the final models.

\subsection{Performance metrics}
Due to the large class imbalance for all outcomes, measuring classification performance with accuracy was inadequate. Rather, we recorded a \textit{confusion matrix} at the end of each batch on the validation set during the learning rate range tests and on the test set during training. The confusion matrix $C$ is a square matrix of dimension $K \times K$ where $K$ is the number of categories, and the $(i,j)^{th}$ element $C_{i,j}$ of $C$ indicates how many of the observations known to be in category $i$ were predicted to be in category $j$. From the confusion matrix, we computed precision, recall and F1-score for each class. Precision, respectively recall, for category $i$, was computed by dividing $C_{i,i}$ (the number of correct predictions for category $i$) by the sum over the $i^{th}$ column of $C$ (the number of predictions made for category $i$), respectively by the sum over the $i^{th}$ row of $C$ (the number of observations in category $i$).

\vspace{-0.5cm}

\begin{equation}
\textrm{precision} = \frac{C_{i,i}}{\sum_{j=1}^{K}C_{j,i}} \hspace{2cm}
\textrm{recall} = \frac{C_{i,i}}{\sum_{j=1}^{K}C_{i,j}}
\end{equation}

\noindent Finally, the F1-score was computed as the harmonic mean of precision and recall:

\begin{align}
\text{F1} & = 2 \times \frac{\text{precision} \times \text{recall}}{\text{precision} + \text{recall}} 
\end{align}

\section{Results}\label{sec:res}

\subsection{Quantitative results}

Classification performance for each prediction task is shown in Tables \ref{table:res_ml_case_cat} to \ref{table:res_ml_acc_cat}. These tables, one for each outcome type, give the precision, recall and F1 score for each algorithm considered, with a random baseline also provided. Each column provides the performance for an individual outcome class with the final column providing the mean value. The best F1 performance for each class is in \textbf{bold text}. 

The three models, XGB, RF, and SVM are compared, together with a random classification baseline. Overall, all three models perform comparably, and all are largely better than the random baseline. Moreover, performance is high, meaning that by using attributes that are only observable before incident occurrence, it is possible to \textit{predict well} four categories of safety outcomes independently extracted by human annotators. This validates the NLP + ML approach of \cite{tixier2016application} for safety-outcome prediction. This also shows that the NLP tool of \cite{tixier2016automated} is robust enough to be used outside of the domains in which it was initially developed and tested: oil and gas versus industrial, energy, infrastructure, and mining.

Of particular note, the \texttt{severity} outcome is predicted by all models with skill much better than random. This is a major improvement over \cite{tixier2016application}, where \texttt{severity} was found to be unpredictable from attributes only.\\

\noindent \textbf{Model stacking}. For \texttt{incident\_type} and \texttt{bodypart}, using the logistic regression model on top of the XGB and RF predictions reaches best performance. For \texttt{severity}, stacking the XGB and RF models also yields a performance boost compared to using XGB or RF alone, but does not outperform the SVM model. This indicates nevertheless the benefit of model stacking, as it improved results for three outcomes out of four. For \texttt{injury\_type}, the SVM model reaches again best performance, and model stacking is slightly worse than using XGB alone.

\begin{table}[H]
\centering
\scalebox{0.6}{
\begin{tabular}{rrccccccc}
 & & access & dropped & eq./tools & PPE & rules & slips/trips/falls & mean \\ 
  \hline
  & prec & 30.81 & 46.71 & 36.50 & 47.93 & 56.98 & 35.95 & 42.48 \\ 
XGB & rec & 38.26 & 35.54 & 54.17 & 59.43 & 25.05 & 56.94 & 44.90 \\ 
  & F1 & \textbf{34.13} & 40.37 & 43.61 & 53.07 & \textbf{34.81} & 44.07 & 41.67 \\ 
\hline
  & prec & 30.71 & 48.35 & 34.12 & 47.07 & 56.91 & 37.51 & 42.45 \\ 
RF & rec & 37.61 & 33.55 & 55.92 & 60.06 & 24.82 & 56.91 & 44.81 \\ 
  & F1 & 33.81 & 39.61 & 42.38 & 52.77 & 34.57 & 45.22 & 41.39 \\
\hline
 & prec & 28.56 & 40.94 & 44.07 & 46.96 & 53.17 & 37.61 & 41.88 \\ 
SVM & rec & 39.00 & 40.85 & 52.25 & 65.06 & 24.46 & 56.18 & 46.30 \\ 
 & F1 & 32.97 & 40.89 & \textbf{47.81} & \textbf{54.55} & 33.51 & 45.06 & 42.46 \\
\hline
 & prec & 29.12 & 41.41 & 42.71 & 45.43 & 56.91 & 37.86 & 42.24 \\ 
XGB+RF & rec & 40.28 & 41.61 & 52.87 & 66.14 & 24.91 & 56.88 & 47.12 \\ 
 & F1 & 33.80 & \textbf{41.51} & 47.25 & 53.87 & 34.65 & \textbf{45.46} & \textbf{42.76} \\ 
\hline
\hline
 & prec & 16.95 & 16.82 & 16.61 & 17.07 & 16.64 & 15.67 & 16.63 \\ 
Random & rec & 11.98 & 9.29 & 30.89 & 10.44 & 15.23 & 21.08 & 16.48 \\ 
 & F1 & 14.04 & 11.97 & 21.60 & 12.95 & 15.90 & 17.98 & 15.74 \\ 
\end{tabular}
}
\captionsetup{size=footnotesize}
\caption[ML: \texttt{incident\_type} test set performance]{Test set performance for \texttt{incident\_type}. Best F1 score per column in \textbf{bold}. \label{table:res_ml_case_cat}}
\end{table}

\vspace{-0.65cm}

\begin{table}[H]
\centering
\scalebox{0.6}{
\begin{tabular}{rrccccccc}
& & eye & finger & head & hand & lower extr. & upper extr. & mean \\ 
\hline
 & prec & 79.70 & 43.59 & 25.82 & 9.71 & 54.17 & 18.13 & 38.52 \\ 
XGB & rec & 84.71 & 44.27 & 27.64 & 26.19 & 24.79 & 18.79 & 37.73 \\ 
 & F1 & 82.13 & 43.93 & \textbf{26.70} & 14.16 & \textbf{34.01} & 18.45 & 36.56 \\ 
\hline
 & prec & 81.55 & 42.82 & 22.54 & 11.18 & 55.56 & 16.37 & 38.34 \\ 
RF & rec & 81.25 & 44.41 & 26.23 & 29.01 & 24.39 & 19.05 & 37.39 \\ 
 & F1 & 81.40 & 43.60 & 24.24 & 16.14 & 33.90 & 17.61 & 36.15 \\ 
\hline
 & prec & 80.07 & 69.49 & 18.31 & 12.06 & 36.11 & 6.43 & 37.08 \\ 
SVM & rec & 80.07 & 36.72 & 29.10 & 32.28 & 29.55 & 16.42 & 37.36 \\ 
 & F1 & 80.07 & \textbf{48.05} & 22.48 & 17.56 & 32.50 & 9.24 & 34.98 \\
\hline
 & prec & 79.70 & 45.90 & 25.35 & 15.59 & 48.15 & 17.54 & 38.71 \\ 
XGB+RF & rec & 86.75 & 43.98 & 27.84 & 28.49 & 25.12 & 19.87 & 38.67 \\ 
 & F1 & \textbf{83.08} & 44.92 & 26.54 & \textbf{20.15} & 33.02 & \textbf{18.63} & \textbf{37.72} \\
\hline
\hline
 & prec & 14.02 & 16.67 & 16.43 & 17.35 & 17.59 & 13.45 & 15.92 \\ 
Random & rec & 14.73 & 24.44 & 12.59 & 19.93 & 15.08 & 9.16 & 15.99 \\ 
 & F1 & 14.37 & 19.82 & 14.26 & 18.55 & 16.24 & 10.90 & 15.69 \\ 
\end{tabular}
}
\captionsetup{size=footnotesize}
\caption[ML: \texttt{bodypart} test set performance]{Test set performance for \texttt{bodypart}. Best F1 score per column in \textbf{bold}. \label{table:res_ml_inj_bod}}
\end{table}

\vspace{-0.65cm}

\begin{table}[H]
\centering
\scalebox{0.6}{
\begin{tabular}{rrccccc}
 & & contusion & cut/puncture & FOB & pain & mean \\ 
  \hline
 & prec & 59.39 & 45.29 & 82.54 & 32.64 & 54.96 \\ 
XGB & rec & 54.29 & 62.23 & 83.87 & 22.27 & 55.66 \\ 
 & F1 & 56.73 & 52.42 & 83.20 & 26.48 & 54.71 \\ 
\hline
 & prec & 40.85 & 42.93 & 84.66 & 52.08 & 55.13 \\ 
RF & rec & 59.59 & 61.65 & 82.05 & 19.33 & 55.66 \\ 
 & F1 & 48.47 & 50.62 & 83.33 & \textbf{28.20} & 52.65 \\ 
\hline
 & prec & 70.89 & 48.17 & 83.07 & 20.14 & 55.57 \\ 
SVM & rec & 54.41 & 61.13 & 83.07 & 30.21 & 57.20 \\ 
 & F1 & \textbf{61.57} & \textbf{53.88} & 83.07 & 24.17 & \textbf{55.67} \\ 
\hline
 & prec & 61.27 & 47.12 & 83.07 & 27.08 & 54.64 \\ 
XGB+RF & rec & 53.27 & 61.43 & 83.96 & 22.81 & 55.37 \\ 
 & F1 & 56.99 & 53.33 & \textbf{83.51} & 24.76 & 54.65 \\ 
\hline
\hline
 & prec & 20.89 & 26.44 & 21.16 & 22.22 & 22.68 \\ 
Random & rec & 34.10 & 35.94 & 13.51 & 10.56 & 23.53 \\ 
 & F1 & 25.91 & 30.47 & 16.49 & 14.32 & 21.80 \\
\end{tabular}
} 
\captionsetup{size=footnotesize}
\caption[ML: \texttt{injury\_type} test set performance]{Test set performance for \texttt{injury\_type}. Best F1 score per column in \textbf{bold}. \label{table:res_ml_inj_type}}
\end{table}

\vspace{-0.65cm}

\begin{table}[H]
\centering
\scalebox{0.6}{
\begin{tabular}{rrccc}
 & & 1st aid & med./restr. & mean \\ 
\hline
 & prec & 74.18 & 53.47 & 63.83 \\ 
XGB & rec & 90.13 & 26.55 & 58.34 \\ 
 & F1 & 81.38 & 35.48 & 58.43 \\ 
\hline
 & prec & 81.39 & 39.93 & 60.66 \\ 
RF & rec & 88.59 & 27.25 & 57.92 \\ 
 & F1 & \textbf{84.84} & 32.39 & 58.62 \\ 
\hline
 & prec & 73.15 & 57.64 & 65.40 \\ 
SVM & rec & 90.82 & 27.26 & 59.04 \\ 
 & F1 & 81.03 & \textbf{37.01} & \textbf{59.02} \\
\hline
 & prec & 73.39 & 56.60 & 65.00 \\ 
XGB+RF & rec & 90.64 & 27.08 & 58.86 \\ 
 & F1 & 81.11 & 36.63 & 58.87 \\ 
\hline
\hline
 & prec & 53.03 & 48.26 & 50.65 \\ 
Random & rec & 85.45 & 15.21 & 50.33 \\ 
 & F1 & 65.44 & 23.13 & 44.29 \\ 
\end{tabular}
}
\captionsetup{size=footnotesize}
\caption[ML: \texttt{severity} test set performance]{Test set performance for \texttt{severity}. Best F1 score per column in \textbf{bold}. \label{table:res_ml_acc_cat}}
\end{table}

\subsection{Qualitative results: attribute importance scores}\label{subsec:attimp}
\begin{figure}[H]
\centering
\includegraphics[width=1\textwidth]{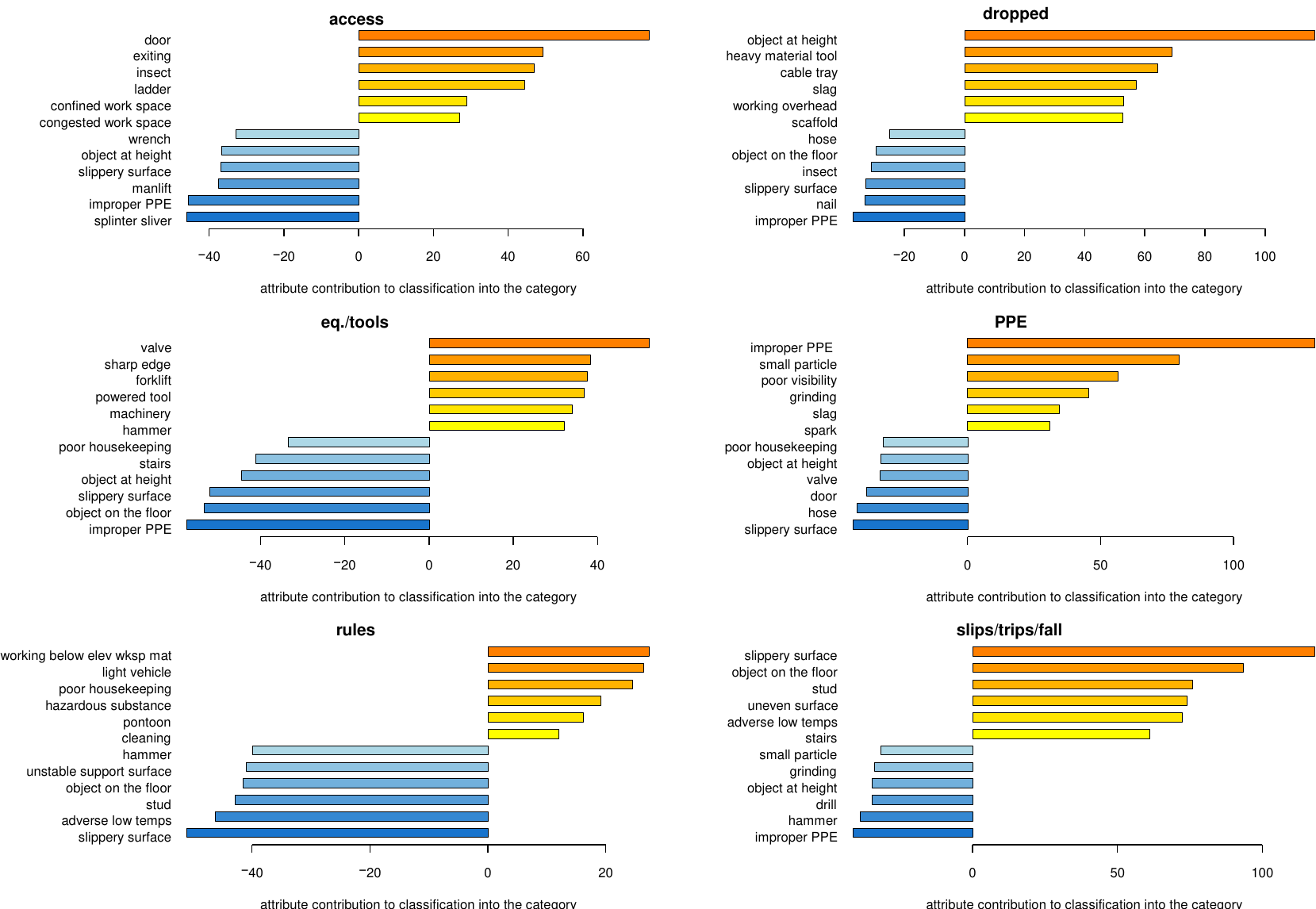}
\captionsetup{size=footnotesize}
\caption{Per-category attribute contribution for \texttt{incident\_type}. \label{fig:attimp_casecat}}
\end{figure}

\begin{figure}[H]
\centering
\includegraphics[width=0.75\textwidth]{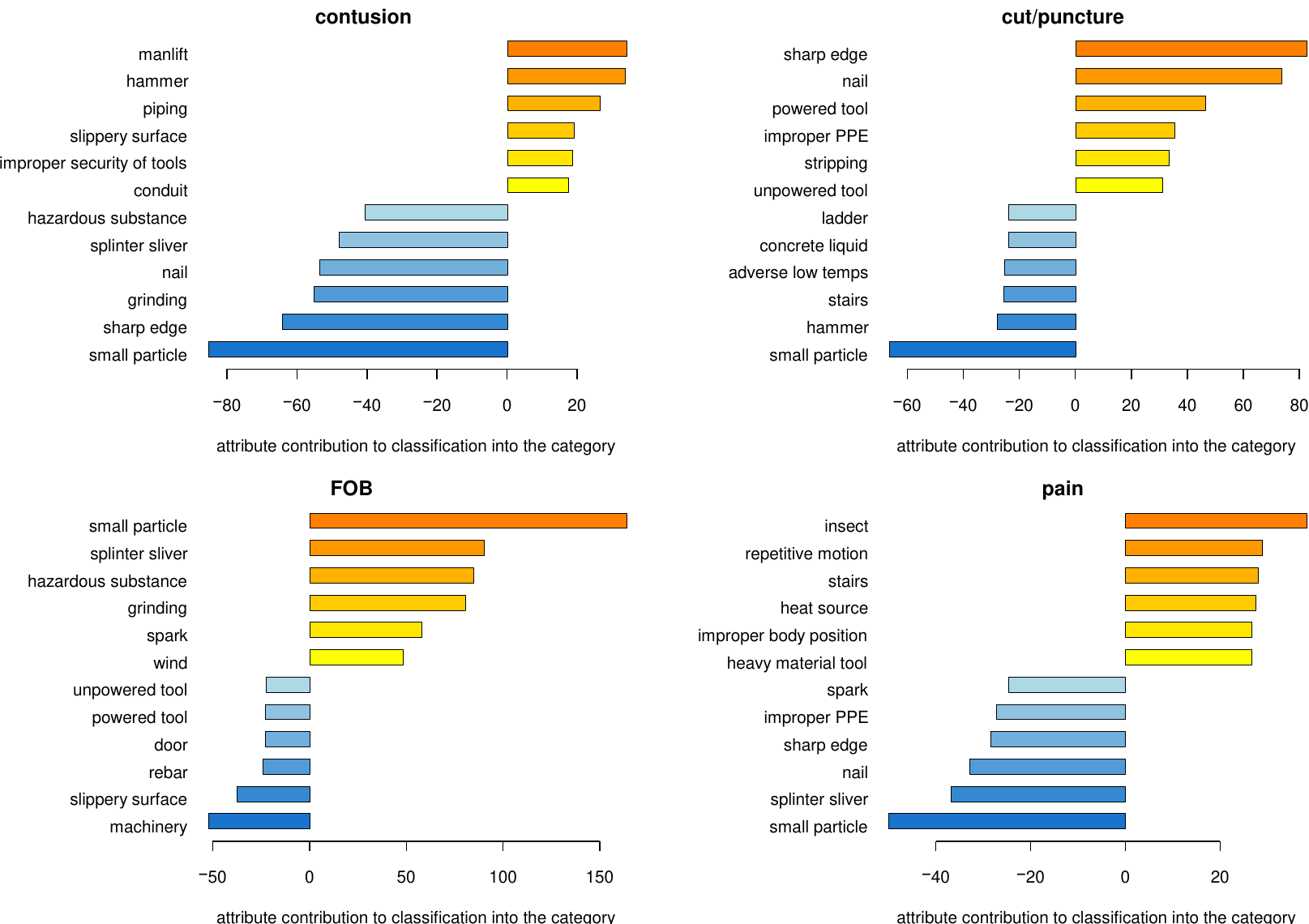}
\captionsetup{size=footnotesize}
\caption{Per-category attribute contribution for \texttt{injury\_type}. \label{fig:attimp_injtype}}
\end{figure}

\begin{figure}[H]
\centering
\includegraphics[width=1\textwidth]{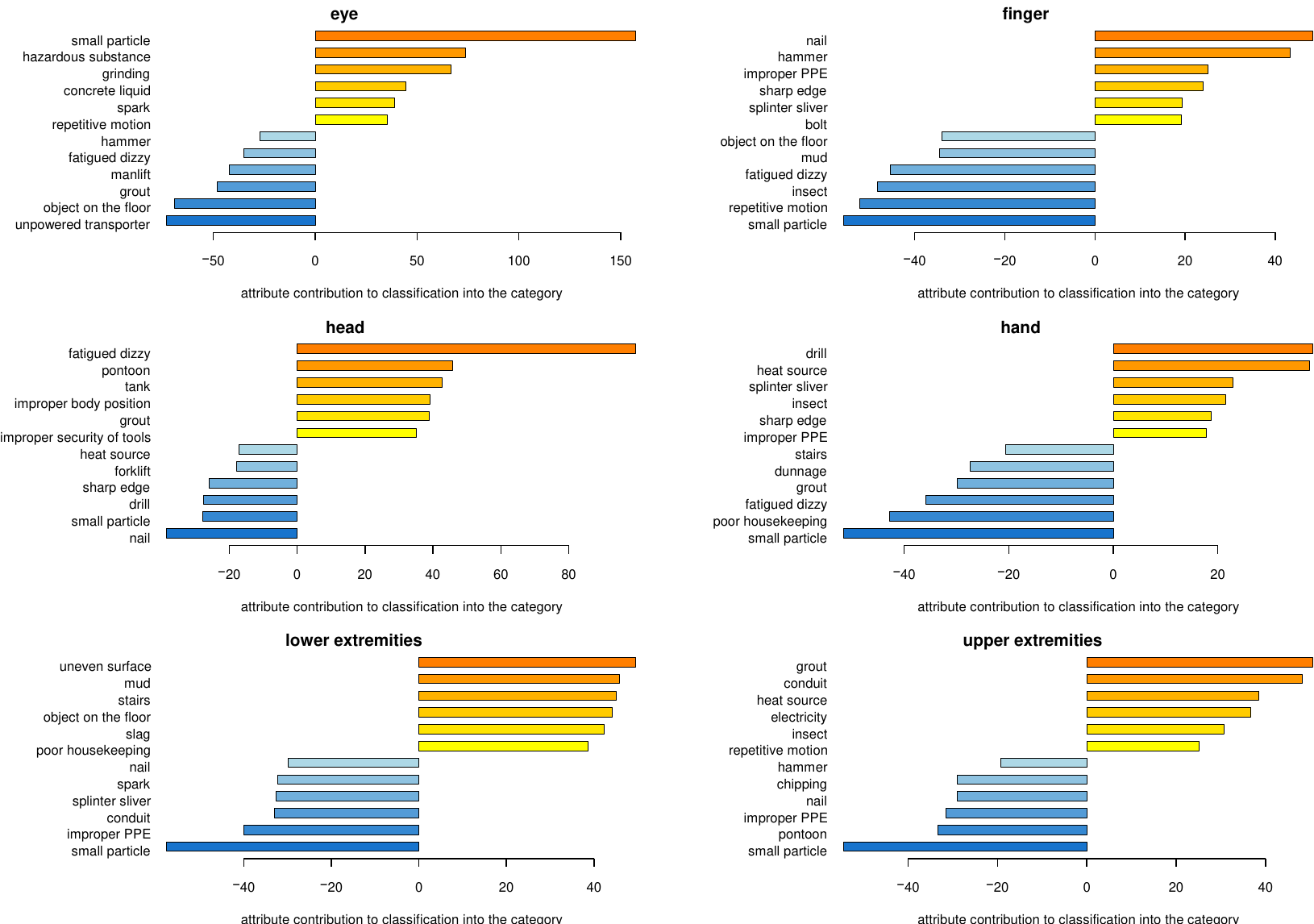}
\captionsetup{size=footnotesize}
\caption{Per-category attribute contribution for \texttt{bodypart}. \label{fig:attimp_body}}
\end{figure}

\begin{figure}[H]
\centering
\includegraphics[width=0.6\textwidth]{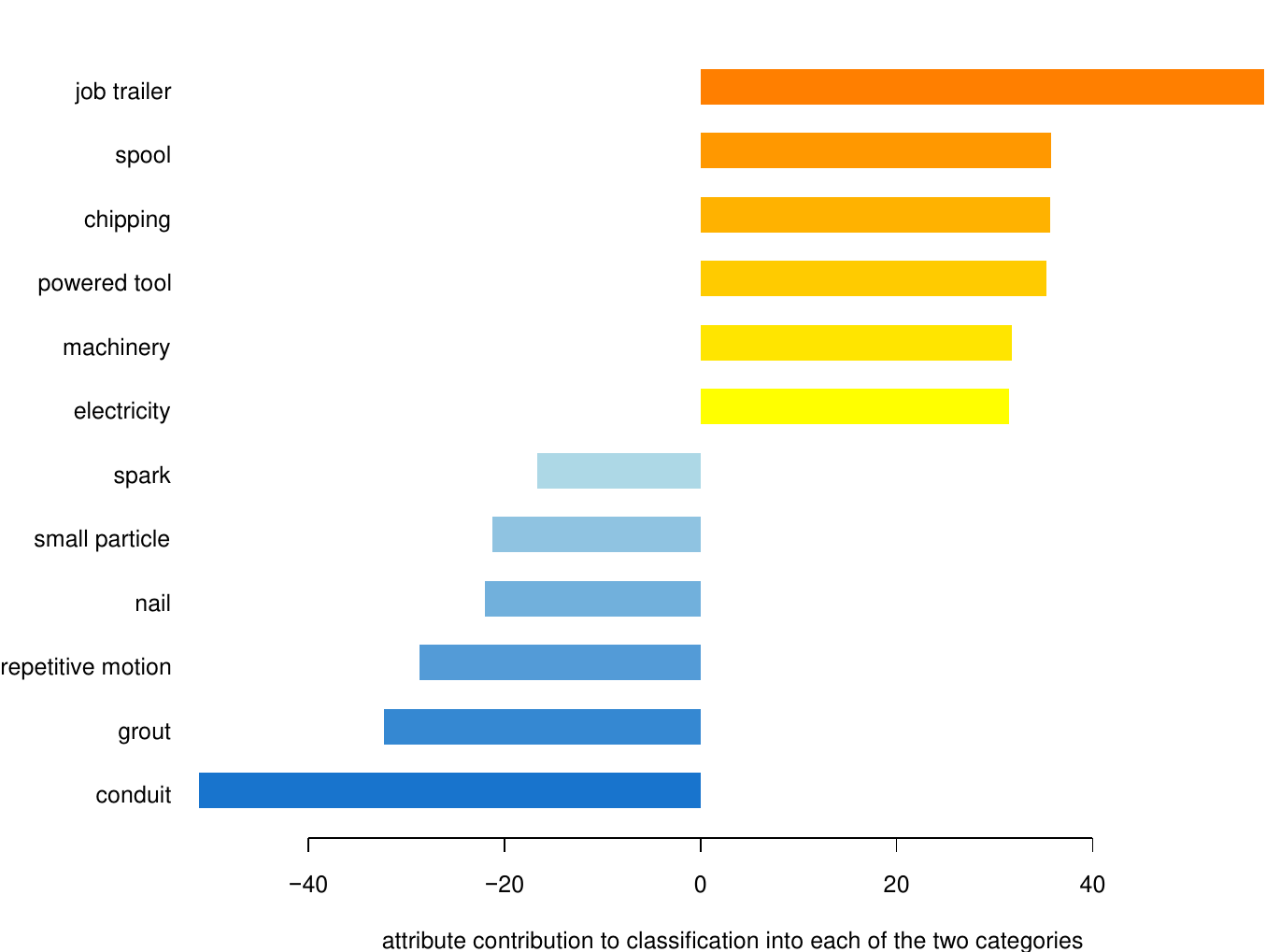}
\captionsetup{size=footnotesize}
\caption{Per-category attribute contribution for \texttt{severity}. \label{fig:attimp_acccat}}
\end{figure}

We selected the linear SVM as our method of choice for computing attribute importance scores. This decision was made for two reasons: the one-vs-rest approach allows the linear SVM to offer attribute importance measures at the categorical level (unlike Random Forest and XGBoost), and these measures are easily interpretable, as was explained in subsection \ref{subsub:attimp},  Results can be seen in the barplots of Figs. \ref{fig:attimp_casecat} to \ref{fig:attimp_acccat}.

More precisely, for each category of each outcome, the barplots show the 6 attributes which are the most predictive of belonging to that category, and the 6 attributes which are the least predictive. 
For the \texttt{severity} outcome, which only contains two categories (\texttt{1st aid} and \texttt{med./restr.}), the top 6 attributes are the ones most predictive of belonging to the \texttt{med./restr.} and respectively the bottom 6 attributes are the ones most predictive of belonging to the \texttt{1st aid} category.

Observing these results, intuitively many of them make sense. For instance, the attributes most predictive of the \texttt{slips/trips/falls} category of the \linebreak \texttt{incident\_type} outcome include `slippery surface', `object on the floor', `uneven surface', `adverse low temperatures', and `stairs'. An example in the \linebreak
\texttt{injury\_type} outcome, the \texttt{cut/puncture} category has `sharp edge', `nail', and `powered tool' as the most predictive attributes.

While finding intuitively important attributes is necessary for model validation and empirical confirmation of our own assumptions, much more interesting for construction safety professionals are those attributes which hold unexpected positions in terms of importance. For instance, `improper PPE' is of low importance for 4 out of the 6 \texttt{incident\_type} categories. A finding which could indicate that the proper PPE is not providing protection against these types of incident and spark further investigation. This is again highlighted in the \texttt{bodypart} outcome where `improper PPE' is ranked of low importance for the upper and lower extremities categories. Arms and legs are arguably the areas least protected by PPE.

Another interesting observation is that for the \texttt{severity} outcome (see Fig. \ref{fig:attimp_acccat}), the most predictive attributes of the medical case category are high-energy attributes (e.g., `powered tool', `machinery', `electricity'), while for the first aid level, the top attributes are associated with low energy tasks (e.g., `repetitive motion', `nail', `small particles'). 
This observation is in accordance with the assumption that the amount of energy in the environment is ultimately what governs injury severity \cite{hallowell2017energy}.

\section{Academic Contributions and Application to Industry}

A major contribution of our study is the validation of the NLP + machine learning pipeline on independent outcomes. The quantitative performance of the prediction methods are high, meaning that by using attributes that are only observable before incident occurrence, it is possible to predict well four categories of safety outcomes \textit{independently} extracted by human annotators. This validates the NLP + ML approach of \cite{tixier2016application} for safety outcome prediction. This also shows that the NLP tool of \cite{tixier2016automated} is robust enough to be used outside of the domains in which it was initially developed and tested: oil and gas versus industrial, energy, infrastructure, and mining.

Most organizations have an injury report database that remains largely unused in terms of advanced analytics. The approach presented in this study allows them to leverage these data. However, while attributes are extracted from raw text, training predictive models requires dependent variables. We therefore recommend construction companies to systematically record as much information as possible whenever a textual report is entered in the database. Also, attribute information could be directly entered by human annotators via check boxes during the injury report procedure. This structured capture of the injury attributes would allow direct application of the findings contained in this paper, independent of the rule-based NLP attribute extraction, so that industry can immediately begin to exploit the mappings discovered between the virtually infinite attribute combinations and outcomes for their own injury data.

Industry practitioners who aim to make safety predictions during work planning may find this method to be highly beneficial. At present, work teams typically identify tasks, consider the work environment, forecast hazards, and devise controls in their job hazard analyses. These tasks are generally performed using the collective experience of the group. However, research has shown that hazard recognition skills are relatively poor in construction with over half of hazards being unidentified prior to work \cite{albert2017empirical}. The attributes described here, however, are neutral, objective descriptors of the means, methods, and environment of the work. Identifying them is much simpler than hazards, because it does not require any thinking or guessing. Indeed, without any specific training or knowledge, the user can list them from a quick analysis of the tools/equipment used to carry out a task (e.g., drilling), and of the environment in which the task is performed (e.g, scaffold, wind). For example, some other examples of attributes include unpowered tool, piping, scaffold, light vehicle, machinery, etc. Additionally, human forecasts of injury likelihood are known to be poor. Our predictive models could be used to augment human judgment during work planning. This method may have broader implications even to design and advanced work packaging.

Other academic contributions, which all represent significant improvements over \cite{tixier2016application}, include (1) the use of two new machine learning models, linear SVM and XGBoost, (2) the use of model stacking, (3) the presentation of a methodology to compute and interpret per-level attribute importance scores, and last but definitely not least, (4) the demonstration that injury severity can be predicted with high skill (it was before thought to be random, or at least not predictable from attributes only).  

\section{Limitations and Recommendations}\label{sec:lims}

Attributes cannot represent everything about the work environment. While the set of 80 attributes we used in this study is well-developed and applicable within the general construction space, more nuanced companies and tasks may require adjustments and/or the creation of new attributes, which represents a time and resource investment. It should be noted that this investment could be offset as the information could be collected at the time of reporting via check boxes, as described in the previous section, rather than relying on the textual description to contain attribute information.

This approach gives a prediction of the outcome of an incident \textit{given that} an incident has occurred. 
Such information is probably more helpful than just knowing the likelihood of incident occurrence.
Indeed, knowing, e.g., that a specific accident type or body part injured are to be expected, allows the adequate preventive measures to be implemented. So, workers are protected regardless of whether or not an accident does occur in the end.
Yet, predicting incident occurrence (i.e., success or failure) is an interesting area of research and should not be neglected.
Training models to make such predictions would require access to \textit{non-injury} reports, i.e., success cases. Reports of near-misses or hazard observations are not ideal for this purpose as they are still associated with some safety issues. What is really needed are descriptions of construction situations where everything went well (non-safety events). Such information may be found, for example, within site diaries or plan of the day documents. Companies could also organize recording campaigns, where annotators would enter their observations of the site directly in terms of attributes. 

The field of machine learning is rapidly evolving. One advantage of the two-step NLP + ML approach is its flexibility: once attributes have been extracted, any machine learning algorithm can be used. Therefore, in the future, the approach can benefit from theoretical machine learning advancements, such as more powerful predictive algorithms.

Future researchers should also test the extent to which this method enables improved human decision-making. In particular, the method could be used during work planning, addressing the limitations of judgment-based bias with empirical data.

\section{Conclusion}
In this paper, we validated the NLP + ML approach of \cite{tixier2016application} by showing that attributes still have high predictive power when the safety outcomes are external and independent. Also, unlike \cite{tixier2016application}, even \textit{injury severity} was well predicted.
Other improvements over \cite{tixier2016application} included: (1) the use of a much larger dataset, (2) two new state-of-the-art models (XGBoost and linear SVM), and (3) model stacking; the adoption of a (4) more straightforward experimental setup with more appropriate performance metrics; and an (5) analysis of per-category attribute importance scores.
Results also showed that the NLP tool of \cite{tixier2016automated} can perform well outside of its original domain.
The proposed approach can be used to complement the current methods used in the construction safety community, which are dominated by opinion-based judgments, perceptions, and risk assessments.

\section{Acknowledgements} 
We thank the anonymous reviewers for their helpful feedback.

\section{Author Contributions} 
Antoine Tixier designed the study, curated the dataset, implemented the models, ran the experiments, interpreted and visualized the results, and wrote the core of the paper.
Henrietta Baker wrote the background and academic contributions sections, reviewed the paper, and led the paper submission process. Matthew Hallowell reviewed the paper.




\section{References}
\small

\bibliographystyle{elsarticle-num} 
\bibliography{bib}

\end{document}